\documentclass[11pt,a4paper]{article}
\usepackage[hyperref]{acl2021}
\usepackage{times}
\usepackage{latexsym}

\usepackage{microtype}

\usepackage{subcaption}
\usepackage{siunitx}
\usepackage{booktabs}
\usepackage{dcolumn}
\usepackage{url}
\usepackage{latexsym}
\usepackage{amsmath}
\usepackage{booktabs}
\usepackage{graphicx}

\usepackage{mathtools}
\usepackage{amsmath}
\usepackage{bm}
\usepackage{listings} 
\usepackage{color}
\usepackage{multirow}
\usepackage[normalem]{ulem}
\usepackage{float}
\newcolumntype{d}[1]{D{.}{.}{#1}}

\aclfinalcopy 

\title{Generalising Multilingual Concept-to-Text NLG \\ with Language Agnostic Delexicalisation}
\author{Giulio Zhou 
  \\Huawei Noah's Ark Lab\\ London, UK\\ \texttt{giuliozhou@huawei.com} \\\And
  Gerasimos Lampouras 
  \\Huawei Noah's Ark Lab\\ London, UK\\ \texttt{gerasimos.lampouras@huawei.com } \\}
  
\date{}

\begin{document}
\maketitle
\begin{abstract}
Concept-to-text Natural Language Generation is the task of expressing an input meaning representation in natural language. Previous approaches in this task have been able to generalise to rare or unseen instances by relying on a delexicalisation of the input. However, this often requires that the input appears verbatim in the output text. This poses challenges in multilingual settings, where the task expands to generate the output text in multiple languages given the same input. In this paper, we explore the application of multilingual models in concept-to-text and propose Language Agnostic Delexicalisation, a novel delexicalisation method that uses multilingual pretrained embeddings, and employs a character-level post-editing model to inflect words in their correct form during relexicalisation. Our experiments across five datasets and five languages show that multilingual models outperform monolingual models in concept-to-text and that our framework outperforms previous approaches, especially in low resource conditions.
\end{abstract}

\section{Introduction}  \label{sec:intro}

\begin{figure}[t]
  \includegraphics[width=\linewidth]{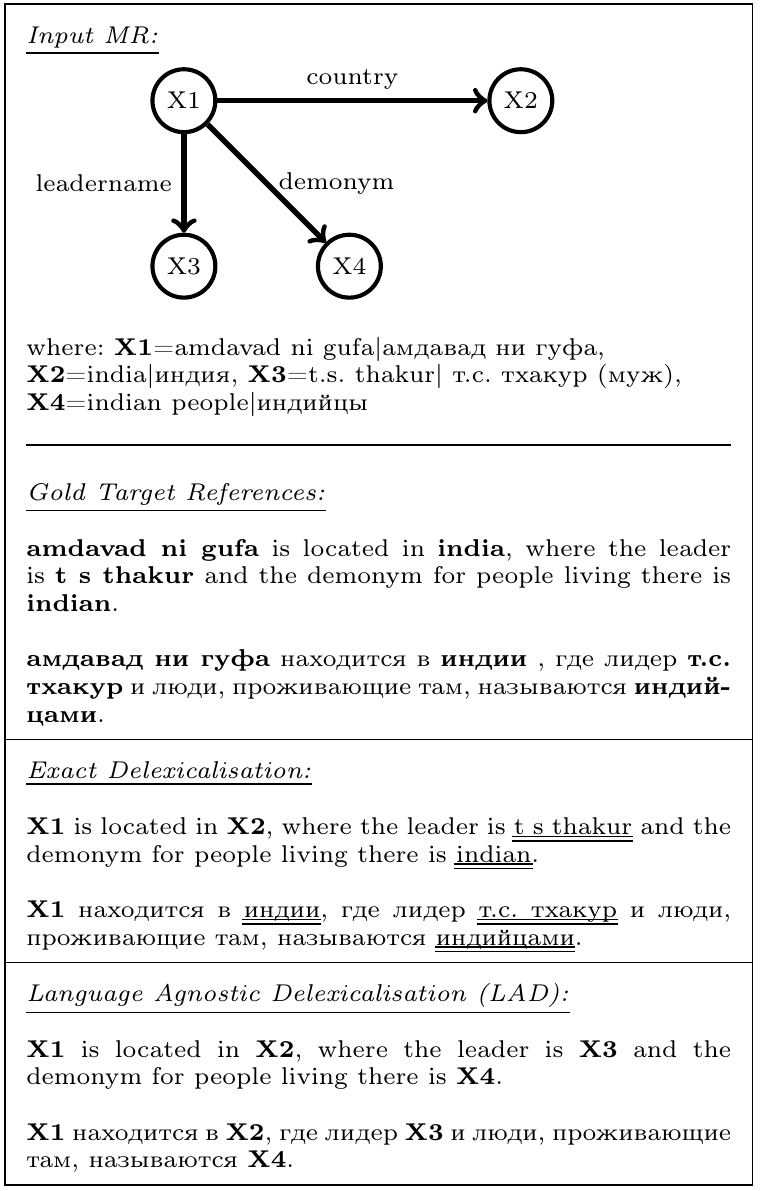}
    \caption{Delexicalisation on WebNLG Challenge 2020 with target output in English and Russian. Double underlining marks text missed by delexicalisation.}
    \label{tab:input_examples}
\vspace{-10pt}
\end{figure}  

Recently, neural approaches to language generation have become predominant in various tasks such as concept-to-text Natural Language Generation (NLG), Summarisation, and Machine Translation thanks to their ability to achieve state-of-the-art performance through end-to-end training \cite{dusek-etal-2018-findings, summary, barrault-etal-2019-findings}. Specifically in Machine Translation, deep learning models have proven easy to adapt to multilingual output \cite{johnson-etal-2017-googles} 
and have been demonstated to successfully transfer knowledge between languages, benefiting both the low and high resource languages  \cite{dabre2020comprehensive}. 

In the concept-to-text NLG task, the language generation model has to produce a text that is an accurate realisation of the abstract semantic information given in the input (Meaning Representation, MR; see Figure~\ref{tab:input_examples}). It is common practice to perform a \textit{delexicalisation} \cite{wen-etal-2015-semantically} of the MR, in order to facilitate the NLG model's generalisation to rare and unseen input; lack of generalisation is a main drawback of neural models \cite{goyal-etal-2016-natural} but is particularly prominent in concept-to-text. 
Delexicalisation consists of a preprocessing and a postprocessing step. In preprocessing, all occurrences of MR values in the text are replaced with placeholders. This way the model learns to generate text that is abstracted away from actual values. In postprocessing (relexicalisation), placeholders are re-filled with values. 

\begin{figure}[htbp]
\centering
  \includegraphics[width=0.96\linewidth]{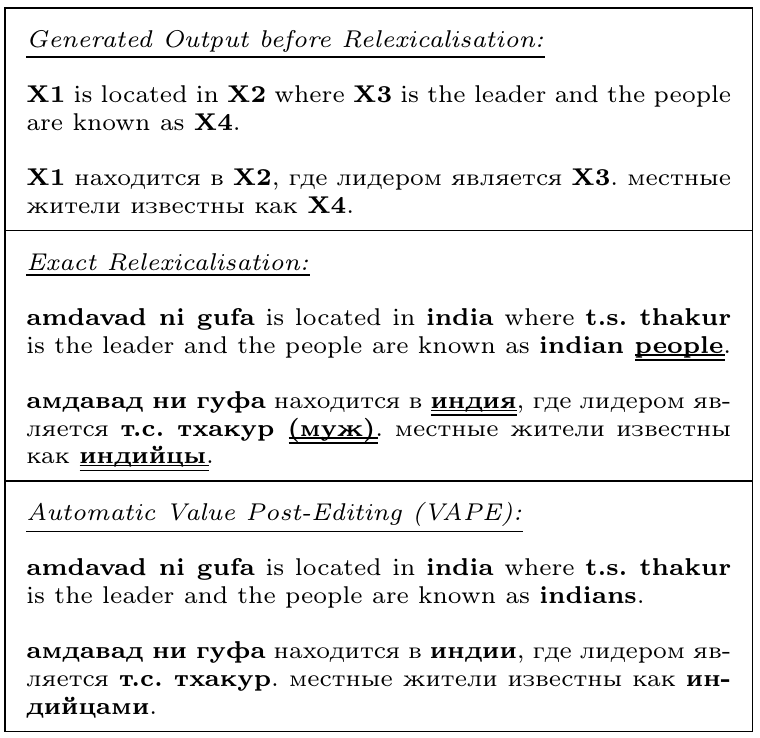}
    \caption{Relexicalisation examples; double underlining marks errors that ignore context.}
    \label{tab:relex_examples}
\vspace{-10pt}
\end{figure}

The main shortcoming of delexicalisation is that its efficacy is bounded by the number of values that are correctly identified. In fact, a naive implementation of ``exact'' delexicalisation (see Figure~\ref{tab:input_examples}) requires the values provided by the MR to appear verbatim in the text, which is often not the case. This shortcoming is more prominent when expanding concept-to-text to the multilingual setting, as MR values in the target language are often only partially provided. Additionally, MR values are usually in their base form, which makes it harder to find them verbatim in text of morphologically rich languages. Finally, relexicalisation also remains a naive process (see Figure~\ref{tab:relex_examples}) that ignores how context should effect the morphology of the MR value when it is added to the text \cite{goyal-etal-2016-natural}.

We propose Language Agnostic Delexicalisation (LAD), a novel delexicalisation method that aims to identify and delexicalise values in the text independently of the language. LAD expands over previous delexicalisation methods and maps input values to the most similar n-grams in the text, by focusing on semantic similarity, instead of lexical similarity, over a language independent embedding space. This is achieved by relying on pretrained multilingual embeddings, e.g. LASER \cite{artetxe-schwenk-2019-massively}. In addition, when relexicalising the placeholders, the values are processed with a character-level post editing model that modifies them to fit their context. Specifically in morphologically rich languages, this post editing results in the value exhibiting correct inflection for its context.

Our goal is to explore the application of multilingual models 
with a focus on their generalisation capability to rare or unseen inputs. 
In this paper, we (i) apply multilingual models and show that they outperform monolingual models in concept-to-text, especially in low resource conditions; (ii) propose LAD and show that it achieves state-of-the-art results, especially on unseen input; (iii) provide experimental analysis across 5 datasets and 5 languages over models with and without pre-training.


\section{Related Work}  \label{sec:related}

Multilingual generation techniques have mostly been the focus of Machine Translation (MT) as the appropriate data (multilingual parallel source and target sentences) are more readily available there. 
Earlier research enabled multilingual generation
with no and partial parameter sharing \cite{luong2015multi, firat-etal-2016-multi}, while \citet{johnson-etal-2017-googles} explored many-to-many translation with full parameter sharing in a universal encoder-decoder framework. Despite the successes of this many-to-many framework, the improvements were mainly attributed to the model's multilingual input. \citet{wang-etal-2018-three} improved on one-to-many translation (i.e. the input is always on a single language, while the output is on many) by introducing special label initialisation, language-dependent positional embeddings and a new parameter-sharing mechanism. 


In other language generation tasks, the vast majority of datasets are only available with English output. To enable output in a different language, a number of Zero-Shot methods have been proposed with the most common practice being to directly use an MT model to translate the output into the target language \cite{wan-etal-2010-cross, shen2018zero, duan-etal-2019-zero}. The MT model can be fine-tuned on task-specific data when those are available \cite{miculicich-etal-2019-selecting}. For the purposes of this paper, we do not consider these previous works as multilingual, as the language generation model is disjoint from the multilingual component, i.e. the pipelined MT model. Contrary to this, \citet{chi2020cross} proposed a cross-lingual pretrained masked language model to generate in multiple languages, outperforming pipeline models on Question Generation and Abstractive Summarisation.

An adaptation of \citet{puduppully-etal-2019-university} was applied to multilingual concept-to-text NLG and participated in the Document-Level Generation and Translation shared task \cite[DGT]{hayashi-etal-2019-findings}. However, this shared task, and in extension the dataset and participating systems, heavily focus on content selection and document generation. Additionally, the input's attributes are constant across train and testing, so there are no unseen data and no need to improve on the model's generalisation capability. 
As the goal of this paper is multilinguality (content selection is a language agnostic task) and generalisation, we opt to not use this dataset.

Multilinguality has also been explored in the related tasks of Morphological Inflection and Surface Realisation in SIGMORPHON \cite{mccarthy-etal-2019-sigmorphon} and MSR \cite{mille-etal-2020-third} challenges. However, our Automatic Value Post-Editing approach focuses mostly on adapting values to context and does not assume additional input such as dependency trees, PoS tags or morphological information that Surface Realisation and Morphological Inflection often requires.


Particularly for concept-to-text NLG, notable previous works includes the approach of \citet{fan-gardent-2020-multilingual} who make use of pretrained language models through the Transformer architecture for AMR-to-text generation in multiple languages, and the WebNLG Challenge 2020 \cite{castro-ferreira-etal-2020-2020}. The goal of WebNLG 2020 was to generate output in both English and Russian but most of the participants focused on monolingual rather than multilingual approaches.

\section{Rare and Unseen Inputs in NLG} \label{sec:rare}


Due to the existence of open-set and numerical attributes in the aforementioned datasets, it is common during testing for MRs to contain rare or unseen values. 
Certain datasets are even more challenging in this regard (e.g. WebNLG Challenge 2020) as they also contain unseen relations in the development and test subsets.
Several techniques have been proposed to mitigate this problem.  

\textbf{Delexicalisation}, also known as anonymisation or masking, is a pre/post-processing procedure that attempts to 
mitigate problems with data sparsity. In preprocessing, all values in the MR that appear verbatim in the target sentence are replaced in both input and output with specific placeholders, e.g. ``X-'' followed by the corresponding attribute (e.g. ``X-type'') so that the placeholder still captures relevant semantic information. In Figure~\ref{tab:input_examples} we use numbered placeholders instead, for clarity and space. The model is trained to generate the target text containing these placeholders, which are subsequently replaced with the corresponding true values (i.e. relexicalised) in post-processing. See Figures~\ref{tab:input_examples} and Figure~\ref{tab:relex_examples} for examples; we mark this strategy as \textit{Exact} due to the exact matching of the values with the text. 
To improve delexicalisation accuracy, n-gram matching \cite{trisedya-etal-2018-gtr} has been proposed as an alternative. Thanks to its simplicity and efficacy, delexicalisation is widely used by many systems, including the winning systems of major concept-to-text NLG shared tasks \cite{gardent-etal-2017-webnlg, dusek-etal-2018-findings, castro-ferreira-etal-2020-2020}.  
Mapping the values as such can be sufficient for simple datasets, 
but otherwise, incorrect or incomplete delexicalisation
will lead to inconsistent input and deteriorate performance. 

Lastly, problems may also occur during relexicalisation as it does not take into account the context in which the placeholders are situated and may result in disfluent sentence. For a simplified example, observe how placing the unedited dates in the placeholders leads to disfluent output in Figure~\ref{tab:relex_examples}.

\textbf{Segmentation strategies} are commonly used in Neural Machine Translation to improve the generalisation ability of models. The objective is to break down words into smaller units, reducing the vocabulary and the number of unseen tokens \cite{sennrich-etal-2016-neural}. 
Unfortunately, applying segmentation in concept-to-text NLG, e.g. using Byte-Pair-Encoding (BPE) subword units \cite{gardent-etal-2017-webnlg, zhang2018attention} or using characters as basic units \cite{goyal-etal-2016-natural, agarwal-dymetman-2017-surprisingly, deriu2017end}, underperforms against delexicalisation. Challenges include capturing long dependencies between segmented words, and generating non-existing words.

\textbf{Copy mechanism} is another method to address unseen input, by allowing the decoder of an encoder-decoder model to draw a token directly from the input sequence instead of generating it from the decoder vocabulary \cite{see-etal-2017-get}. 
While applications of the copy mechanism in concept-to-text NLG have achieved overall good results \cite{chen2018general, elder-etal-2018-e2e, gehrmann-etal-2018-end}, when dealing with rare and unseen inputs delexicalisation is still preferable \cite{shimorina-gardent-2018-handling}. To improve the generalisation ability of copy mechanism models, \citet{roberti2019copy} propose applying the copy mechanism to character-level NLG systems. This is combined with an additional optimisation phase during training where the encoder and decoder are switched. 

\section{Language Agnostic Delexicalisation}  \label{sec:lad}

\begin{figure}[t]
\centering
  \includegraphics[width=\linewidth]{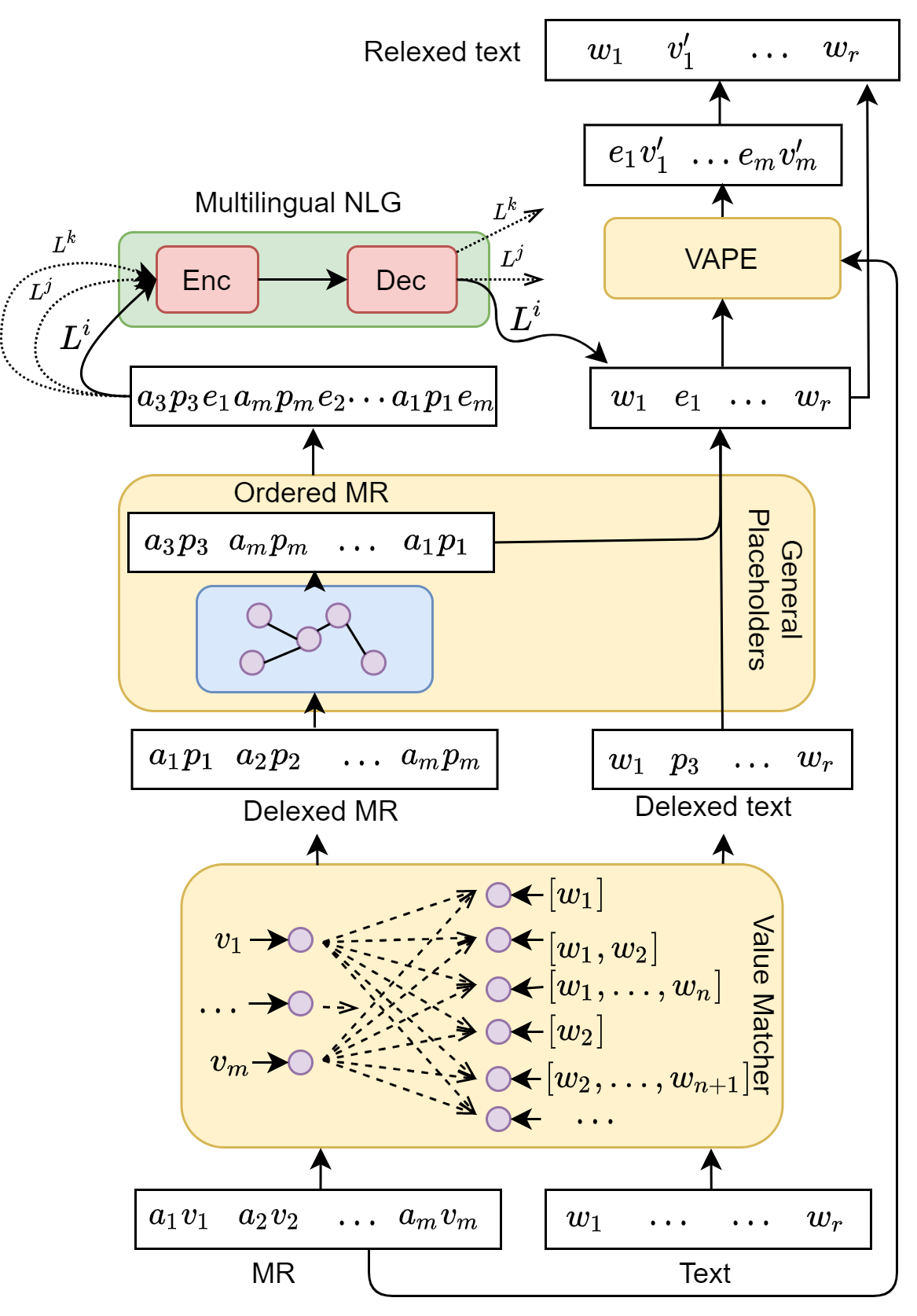}
\caption{Language Agnostic Delexicalisation outline.}
  \label{tab:lad}
\end{figure} 

In order to address the shortcomings of previous approaches to generalise over rare or unseen inputs, 
especially in cases of multilingual output, we propose Language Agnostic Delexicalisation (LAD). 
Figure~\ref{tab:lad} shows an overview of LAD; the input and output are first delexicalised using pretrained language-independent embeddings, and (optionally) ordered.
The multilingual generation model is trained on the delexicalised training data, and the output is relexicalised using automatic value post-editing to ensure that the values fit the context. Each component is described in more detail bellow. 

To enable multilingual generation, we adapt the universal encoder-decoder framework via ``\textit{target forcing}" \cite{johnson-etal-2017-googles} since it can be directly applied to any NLG model without the need to modify the latter's architecture. To do so, we extend the input MR in the encoder with a language token that signals which language the model should generate output in. In addition, we follow \citet{wang-etal-2018-three} and initialise the decoder with the language token. The rest of the components (i.e. delexicalisation, ordering, and value post-editing) are orthogonal to the model's architecture. 

\subsection{Value Matching}\label{sec:valuematching}

As discussed in Section~\ref{sec:rare}, one of the challenges of delexicalisation is matching the MR values with corresponding words in the text, especially in the multilingual setting. 
Even when the MR values are in the same language as the target, we observe from the examples in Figure~\ref{tab:input_examples} that token overlapping methods (i.e. exact and n-gram matching) are not sufficient to generate a complete and accurate delexicalisation as values may appear differently.

To counter this problem, LAD performs matching by mapping MR values to n-grams based on the similarity of their representations. Specifically, it calculates the similarity between a value $v$ and all word n-grams $w_i \dots w_j$ in the text, with $j-i < n$ and $n$ set to the maximum value length observed in the training data. LAD employs LASER \cite{artetxe-schwenk-2019-massively} to generate language agnostic sentence embeddings of the values and n-grams, and calculates their distance via cosine similarity. Given an MR and text, all possible value and n-gram comparisons are calculated and the matches are determined in a greedy fashion. 

\subsection{Generic Placeholders and Ordering}

In Section~\ref{sec:rare}, we discussed how the WebNLG datasets are more challenging because they contain unseen attributes in the development and test subsets, in addition to unseen values. This is problematic when we use attribute-bound placeholders (e.g. ``X-type'') as unseen attributes will result in unseen placeholders. Following \citet{trisedya-etal-2018-gtr}, for the WebNLG datasets, LAD uses numbered generic placeholders ``X\#" (e.g. ``X1''). Unfortunately, the adoption of generic placeholders creates problems for relexicalisation as it becomes unclear which input value should replace which placeholder. We address this by ordering the model's input based on the graph formed by its RDF triples, again by following \citet{trisedya-etal-2018-gtr}. We traverse every edge in the graph, starting from the node with the least incoming edges (or randomly in case of ties) and then visit all nodes via BFS (breadth-first search). We then trust that the model will learn to respect the input order when generating, and follow the order to relexicalise the placeholders. 

We note that this is only required for models that employ delexicalisation strategies and for datasets with unseen attributes (i.e. the WebNLG Challenge datasets). Concept-to-text NLG systems do not generally require ordered input \cite{wen-etal-2015-semantically}. 



\subsection{Automatic Value Post-Editing} \label{sec:vape}

As discussed in Section~\ref{sec:rare}, a naive relexicalisation of the placeholders may lead to disfluent sentences, as the procedure does not take into account the context in which the placeholders have been placed. For example, in the sentence ``there are 2 X that have free parking'', if we need to replace the placeholder ``X" with the MR value ``guesthouse'' , the value should be pluralised to fit the context. This problem is more evident in morphologically rich languages, where more factors affect the value's form.
To alleviate this, the LAD framework incorporates an Automatic Value Post-Editing component, consisting of a character-level seq2seq model that iterates over values as they are placed in the text and modifies them to fit the context of their respective placeholders. \citet{anastasopoulos-neubig-2019-pushing} has already shown the benefits of character models on morphological inflection generation, but no previous work has addressed how relexicalisation should adapt to context.

Our proposed VAPE model requires as input the MR placeholder $e_i$, original value $v_i$ and corresponding NLG output $w'_1\dots w'_n$ for context; these are serialised and passed to the encoder. Similar to the multilingual model, we add an appropriate language token $L$ before the NLG output. The output of VAPE is the MR value $v'_i$ in the proper form.
$$ \{ e_i \; v_i \; \mbox{[SEP]} \; L \; w'_1\dots w'_n \} \rightarrow v'_i$$

The training signal for VAPE is obtained during delexicalisation. For a given delexicalisation strategy, we obtain all pairs of MR values and matching n-grams in the training data, and subsequently train VAPE using these n-grams as the targets. Therefore, the VAPE model is dependent on the quality of the delexicalisation strategy; specifically for exact delexicalisation, VAPE cannot be trained as the MR values and matching n-grams are the same.

Most edits VAPE performs concern incorrect inflections, but it is not limited to morphological edits and has the potential to deal with various types of modifications. During our experiments we observed VAPE performing value re-formatting (e.g. ``1986\_04\_15" $\rightarrow$ \textit{``April 15th 1986}"), synonym generation (e.g. \textit{``east"} $\rightarrow$ \textit{``oriental"}) and value translation (e.g. ``bbc'' from Latin to Cyrillic).



\section{Experiments}


For our experiments we use five datasets and calculate BLEU-4 \cite[$\uparrow$]{papineni-etal-2002-bleu}, METEOR \cite[$\uparrow$]{banerjee-lavie-2005-meteor}, chrF++ \cite[$\uparrow$]{popovic-2015-chrf}, and TER \cite[$\downarrow$]{snover2006study}. 



The WebNLG Challenge 2017 \cite[WebNLG17]{gardent-etal-2017-webnlg} data consists of sets of RDF triple pairs and corresponding English texts in 15 DBPedia categories. 
For our purposes, we will be using a later work \cite{shimorina-etal-2019-creating} that introduced a machine translated Russian version of WebNLG17, a part of which was post edited by humans. Due to the limited amount of human corrected Russian sentences, and to facilitate the most accurate evaluation, we use these solely for testing. To ensure that half of the domains in the new test set remain unseen during training, we create our own train/dev/test split by retaining the following DBPedia categories from training and development sets: Astronaut, Monument and University.

The latest incarnation of the WebNLG Challenge \cite[WebNLG20]{castro-ferreira-etal-2020-2020} is fully human annotated for both English and Russian. We use this as the main dataset in our experiments, as it is designed to promote multilinguality. 
However, due to the fact that the provided test set does not contain unseen Russian instances, we perform our experiment on a custom split (WebNLG20*) ensuring that part of the domains in the test data remain unseen during training. The split was performed similarly to the previously described WebNLG17.  

MultiWOZ 2.1 \cite{eric2020multiwoz} and CrossWOZ \cite{zhu-etal-2020-crosswoz} are datasets of dialogue acts and corresponding utterances in English and Chinese respectively. The two datasets share the same structure, with MultiWOZ covering 7 domains and 25 attributes, and CrossWOZ covering 5 domains and 72 attributes; 4 of the domains are common in both datasets though CrossWOZ has more attached attributes. 
Multilingual WOZ 2.0 \cite{mrksic-etal-2017-neural} is also a dialogue dataset with utterances available in three languages: English, Italian and German. Its scope is more limited than MultiWOZ and CrossWOZ as it only covers a single domain.

For all models in our experiments, the input consists of a simple linearisation of the MRs. Particularly, for the delexicalisation based models, the values are extended with their respective placeholders as shown in the following example: ``ENTITY\_1 meyer werft \textit{location} ENTITY\_2 germany".

\subsection{Ablation Study}\label{sec:ablation}

First we perform an ablation study to determine how the different components of \textit{LAD} (ordering and VAPE) affect its performance; \textit{LAD} being our full Language Agnostic Delexicalisation model as described in Section~\ref{sec:lad}. In addition to \textit{LAD}, where these components are incrementally removed, we explore how their addition would influence exact and n-gram delexicalisation \cite{trisedya-etal-2018-gtr}. We do not explore adding VAPE to exact delexicalisation (there is no \textit{EXACT + O + V} variant), as it cannot be trained in this setting (see Section~\ref{sec:vape}). 

In Table~\ref{tab:ablation}, we observe that both components are beneficial, but less so for seen English data. For the more morphologically rich and lower resourced Russian, the components are helpful for both seen and unseen.
VAPE leads to an improvement in performance in almost all cases and even when added on \textit{NGram}. An exception is unseen English data, where removing VAPE is beneficial; this suggests that VAPE is overeager to make edits in English. 

By studying the output, we observe that VAPE modified 
20\% of values in English, and 66\% in Russian; directly copying the value was insufficient in Russian where proper inflection is needed. 
We identified three consistent errors where copying the original value would be preferable to using VAPE: the removal of date information (e.g. ``1969-09-01 $\rightarrow$ 1st, 1969"), misspelling of proper nouns (e.g. ``atatürk monument" $\rightarrow$ ``atat erk monument"), and mishandling of long values (e.g. ``ottoman army soldiers killed in the battle of baku" $\rightarrow$ ``ottoman army soldiers killed in the batttle of kiled in the bathe batom"). 
We observe that these errors occur more frequently for English unseen cases, but could be reduced by extending VAPE with a control mechanism that decides whether copying the values themselves is preferable. 
Such errors occur in part because VAPE, as a character-level model, suffers from the same challenges as other segmentation methods (see Section~\ref{sec:rare}). However, since VAPE's input is much shorter, the problem is not as prevalent. 
Overall, \textit{LAD} outperforms the previous delexicalisation strategies \textit{Exact} and \textit{NGram}, and VAPE is shown to be integral to its performance.

\begin{table}[!t]
\renewcommand{\arraystretch}{1.1}
\centering
\resizebox{\columnwidth}{!}{\begin{tabular}{|l|c|c|c|c|c|r|}
\hline
\multirow{2}{*}{} & \multicolumn{3}{c|}{English}                     & \multicolumn{3}{c|}{Russian}                     \\ \cline{2-7} 
                  & A              & S              & U              & A              & S              & \multicolumn{1}{c|}{U}               \\ \hline
Exact             & 0.56          & 0.62          & 0.18          & 0.21          & 0.25          &  0.03           \\
Exact + O         & 0.52          & 0.54           & 0.38          & 0.19          & 0.20          & 0.14          \\ \hline
NGram             & 0.56          & 0.62          & 0.16          & 0.22          & 0.27          &  0.04           \\
NGram + O         & 0.53          & 0.55          & \textbf{0.40} & 0.23          & 0.25          & 0.15          \\
NGram + O+V       & 0.54          & 0.57          & 0.33          & 0.31          & 0.35          & 0.16          \\ \hline
LAD - O-V         & 0.59          & 0.65          & 0.18          & 0.23          & 0.28          &  0.03           \\
LAD  - V          & 0.60          & 0.63          & 0.39          & 0.24          & 0.26          & 0.16          \\
LAD               & \textbf{0.62} & \textbf{0.66} & 0.32          & \textbf{0.37} & \textbf{0.42} & \textbf{0.21} \\ \hline
\end{tabular}}
\caption{BLEU on WebNLG20* for delexicalisation models augmented with generic placeholders+ordering, and value post-edit. A = All categories; S = Seen categories; U = Unseen categories; O = generic placeholders+ordering; V = Value post-edit.}
\label{tab:ablation}
\end{table}


\subsection{Monolingual vs Multilingual} \label{sec:monoVmulti}

\begin{table}[!t]
\renewcommand{\arraystretch}{1.1}
\resizebox{\columnwidth}{!}{\begin{tabular}{|c|c|c|c|c|c|c|c|c|}
\hline
\multicolumn{2}{|c|}{\multirow{2}{*}{}} & \multicolumn{3}{c|}{WOZ 2.0}                         & \multicolumn{2}{c|}{\begin{tabular}[c]{@{}c@{}}WebNLG\\ 2020*\end{tabular}} & \multicolumn{2}{c|}{\begin{tabular}[c]{@{}c@{}}MultiWOZ\\ + CrossWOZ\end{tabular}} \\ \cline{3-9} 
\multicolumn{2}{|c|}{}                  & en             & it             & de             & en                                   & ru                                  & en                                       & zh                                      \\ \hline
\multirow{2}{*}{Word}      & Mono       & \textbf{0.66} & 0.56          & 0.59          & \textbf{0.57}                       & 0.31                               & 0.56                                    & \textbf{0.68}                          \\
                           & Multi      & 0.65          & \textbf{0.57} & \textbf{0.61} & \textbf{0.57}                                & \textbf{0.33}                      & \textbf{0.57}                           & 0.66                                 \\ \hline \hline
\multirow{2}{*}{LAD}       & Mono       & 0.66          & 0.58          & 0.56          & 0.58                                & 0.32                               & 0.56                                    & \textbf{0.68}                                   \\
                           & Multi      & \textbf{0.68} & \textbf{0.59} & \textbf{0.57} & \textbf{0.62}                       & \textbf{0.37}                      & \textbf{0.58}                           & \textbf{0.68}                                   \\ \hline
\end{tabular}}
\caption{BLEU for mono- and multilingual models.}
\label{tab:monoVmulti}
\end{table}

Here we explore the performance of monolingual and multilingual models on concept-to-text datasets.
The \textit{Word} model has the exact same architecture as \textit{LAD} but no delexicalisation is performed, and consequently no automatic value post-editing and no ordering. Since there is no relexicalisation that needs to occur during post-processing, the input to the \textit{Word} model needs not be specifically ordered, and is just a concatenation of the RDF triples as they appear in the original dataset. For multilingual, we add the appropriate language tokens on the input of \textit{Word}, in the same manner we added them to \textit{LAD}. 
For the monolingual (\textit{Mono}) configuration we train the models to produce a single language, while for multilingual (\textit{Multi}) we train them to produce all languages available in that dataset. Please refer to Table~\ref{tab:monoVmulti} for the results.

\begin{table*}[!htb]
\renewcommand{\arraystretch}{1.1}
\begin{subtable}{1\textwidth}
\sisetup{table-format=-1.2}   
\centering
\resizebox{\textwidth}{!}{\begin{tabular}{|l|c|c|c|c|c|c|c|c|c|c|c|c|}
\hline
\multirow{2}{*}{English} & \multicolumn{4}{c|}{All Categories}                   & \multicolumn{4}{c|}{Seen Categories}                           & \multicolumn{4}{c|}{Unseen Categories}                         \\ \cline{2-13} 
                         & BLEU           & METEOR        & chrF++        & TER  & BLEU           & METEOR        & chrF++        & TER           & BLEU           & METEOR        & chrF++        & TER           \\ \hline
Word                     & 0.57          & 0.36          & 0.63          & 0.44 & 0.64          & 0.43          & 0.72          & 0.36          & 0.10           & 0.11          & 0.26          & 0.90          \\ \hline
Char                     & 0.54          & 0.35          & 0.52          & 0.47 & 0.61          & 0.41          & 0.70          & 0.40          & 0.05           & 0.09          & 0.24          & 0.89          \\ \hline
BPE                      & 0.54          & 0.35          & 0.63          & 0.49 & 0.62          & 0.42          & 0.72          & 0.42          & 0.07           & 0.09          & 0.24          & 0.97          \\ \hline
SP                       & 0.58          & 0.37          & 0.64          & 0.42 & \textbf{0.66}          & 0.44          & 0.74          & 0.34          & 0.07           & 0.09          & 0.23          & 0.96          \\ \hline
Copy                     & 0.57          & 0.36          & 0.63          & 0.45 & 0.59          & 0.38          & 0.65          & 0.42          & \textbf{0.38} & 0.27          & 0.51          & \textbf{0.61} \\ \hline \hline
LAD                      & \textbf{0.62} & \textbf{0.42} & \textbf{0.71} & \textbf{0.36} & \textbf{0.66} & \textbf{0.45} & \textbf{0.75} & \textbf{0.31} & 0.32          & \textbf{0.30} & \textbf{0.54} & \textbf{0.61} \\ \hline
\end{tabular}}

\end{subtable}

\medskip
\begin{subtable}{1\textwidth}
\sisetup{table-format=1.0} 
\centering
\resizebox{\textwidth}{!}{\begin{tabular}{|l|c|c|c|c|c|c|c|c|c|c|c|c|}
\hline
\multirow{2}{*}{Russian} & \multicolumn{4}{c|}{All Categories}                            & \multicolumn{4}{c|}{Seen Categories}                           & \multicolumn{4}{c|}{Unseen Categories}                         \\ \cline{2-13} 
                         & BLEU           & METEOR        & chrF++        & TER           & BLEU           & METEOR        & chrF++        & TER           & BLEU           & METEOR        & chrF++        & TER           \\ \hline
Word                     & 0.33          & 0.41          & 0.44          & 0.63          & 0.42          & 0.51          & 0.54          & 0.56          & 0.02           & 0.13          & 0.20          & 0.94          \\ \hline
Char                     & 0.30          & 0.41          & 0.44          & 0.67          & 0.38          & 0.50          & 0.53          & 0.61          & 0.01           & 0.13          & 0.20          & 0.91          \\ \hline
BPE                      & 0.25          & 0.38          & 0.44          & 0.71          & 0.32          & 0.48          & 0.54          & 0.65          & 0.02           & 0.11          & 0.19          & 0.97          \\ \hline
SP                       & 0.34          & 0.41          & 0.45          & 0.62          & \textbf{0.44} & 0.54          & 0.56          & 0.53          & 0.01           & 0.10          & 0.18          & 0.98          \\ \hline
Copy                     & 0.24          & 0.35          & 0.39          & 0.74          & 0.29          & 0.42          & 0.46          & 0.72          & 0.02           & 0.14          & 0.20          & 0.91          \\ \hline \hline
LAD                      & \textbf{0.37} & \textbf{0.51} & \textbf{0.55} & \textbf{0.55} & 0.42          & \textbf{0.57} & \textbf{0.60} & \textbf{0.51} & \textbf{0.21} & \textbf{0.34} & \textbf{0.42} & \textbf{0.71} \\ \hline
\end{tabular}}

\end{subtable}
\caption{WebNLG20* results for Multilingual models.}
\label{tab:WebNLG20}
\end{table*}

\begin{table*}[!htb]
\renewcommand{\arraystretch}{1.1}
\begin{subtable}{1\textwidth}
\sisetup{table-format=-1.2}   

\centering
\resizebox{\textwidth}{!}{\begin{tabular}{|l|c|c|c|c|c|c|c|c|c|c|c|c|}
\hline
\multicolumn{1}{|c|}{\multirow{2}{*}{English}} & \multicolumn{4}{c|}{All Categories}                            & \multicolumn{4}{c|}{Seen Categories}                           & \multicolumn{4}{c|}{Unseen Categories}                         \\ \cline{2-13} 
\multicolumn{1}{|c|}{}                         & BLEU           & METEOR        & chrF++        & TER           & BLEU           & METEOR        & chrF++        & TER           & BLEU           & METEOR        & chrF++        & TER           \\ \hline
SP                                             & 0.58          & 0.37          & 0.64          & 0.42          & 0.66          & 0.44          & 0.74          & 0.34          & 0.07           & 0.09          & 0.23          & 0.96          \\ \hline
LAD                                            & 0.62          & 0.42          & 0.71          & 0.36          & 0.66          & 0.45          & \textbf{0.75}          & 0.31          & 0.32          & 0.30          & 0.54          & 0.61          \\ \hline \hline
mBART                                          & 0.66          & 0.44          & 0.74          & 0.33          & 0.67          & 0.45          &\textbf{0.75}          & 0.32          & 0.58          & 0.41          & 0.70          & 0.44          \\ \hline
mB-LAD                                         & 0.66          & 0.44          & 0.74          & \textbf{0.31} & \textbf{0.68} & \textbf{0.46} & \textbf{0.75} & \textbf{0.30} & 0.52          & 0.38          & 0.68          & 0.44          \\ \hline
mB-LAD+                                        & \textbf{0.67} & \textbf{0.45} & \textbf{0.75} & \textbf{0.31} & \textbf{0.68}          & \textbf{0.46} & \textbf{0.75} & \textbf{0.30} & \textbf{0.61} & \textbf{0.42} & \textbf{0.71} & \textbf{0.37} \\ \hline
mB-LAD-SPE                                      & 0.66          & 0.44          & 0.74          & \textbf{0.31} & 0.67          & 0.45          & \textbf{0.75} & \textbf{0.30} & 0.59          & 0.41          & 0.70          & 0.38          \\ \hline
\end{tabular}}

\end{subtable}

\medskip
\begin{subtable}{1\textwidth}
\sisetup{table-format=4.0} 

\centering
\resizebox{\textwidth}{!}{\begin{tabular}{|l|c|c|c|c|c|c|c|c|c|c|c|c|}
\hline
\multicolumn{1}{|c|}{\multirow{2}{*}{Russian}} & \multicolumn{4}{c|}{All Categories}                            & \multicolumn{4}{c|}{Seen Categories}                           & \multicolumn{4}{c|}{Unseen Categories}                         \\ \cline{2-13} 
\multicolumn{1}{|c|}{}                         & BLEU           & METEOR        & chrF++        & TER           & BLEU           & METEOR        & chrF++        & TER           & BLEU           & METEOR        & chrF++        & TER           \\ \hline
SP                                             & 0.34          & 0.41          & 0.45          & 0.62          & 0.44          & 0.54          & 0.56          & 0.53          & 0.01           & 0.10          & 0.18          & 0.98          \\ \hline
LAD                                            & 0.37          & 0.51          & 0.55          & 0.55          & 0.42          & 0.57          & 0.60          & 0.51          & 0.21          & 0.34          & 0.42          & 0.71          \\ \hline \hline
mBART                                          & 0.37          & 0.50          & 0.51          & 0.57          & 0.43          & 0.56          & 0.57          & 0.52          & 0.15          & 0.33          & 0.35          & 0.78          \\ \hline
mB-LAD                                         & 0.41          & 0.54          & 0.58          & 0.51          & \textbf{0.45}          & \textbf{0.58} & \textbf{0.61} & \textbf{0.49} & 0.29          & 0.44          & 0.48          & 0.59          \\ \hline
mB-LAD+                                        & 0.42          & 0.55          & 0.58          & 0.51          & 0.41          & 0.57          & 0.60          & 0.50          & 0.41          & \textbf{0.52} & \textbf{0.55} & 0.52          \\ \hline
mB-LAD-SPE                                      & \textbf{0.46} & \textbf{0.57} & \textbf{0.59} & \textbf{0.47} & 0.42 & 0.57          & 0.60          & \textbf{0.49} & \textbf{0.44} & \textbf{0.52} & \textbf{0.55} & \textbf{0.49} \\ \hline
\end{tabular}}

\end{subtable}

\caption{WebNLG20* results for Pretrained Multilingual models.}
\label{tab:WebNLG20_mbart}
\end{table*}



We observe that the multilingual models outperform their monolingual counterpart in most datasets and languages, especially with \textit{LAD} as its delexicalisation and relexicalisation modules are more robust to multilingual input and output. 
Specifically for the MultiWOZ and CrossWOZ datasets, in the monolingual setting the models are trained exclusively on the respective dataset, i.e. MultiWOZ for English, and CrossWOZ for Chinese. For multilingual, we take advantage of the fact that these datasets share the same structure, and train the models on both datasets. For English, we observe that the multilingual model improves, suggesting that domain knowledge is transferred from CrossWOZ. For Chinese however, the multilingual \textit{Word} model underperforms. This is not very surprising, as the overlap between the datasets is favourable to MultiWOZ, i.e. most of the attributes of MultiWOZ also appear in CrossWOZ, while the majority of CrossWOZ's attributes do not appear in MultiWOZ.

\begin{table*}[!t]
\renewcommand{\arraystretch}{1.1}
\centering
\resizebox{\textwidth}{!}{\begin{tabular}{|l|c|c|c|c|c|c|c|c|c|c|c|c|}
\hline
\multicolumn{1}{|c|}{\multirow{2}{*}{English}} & \multicolumn{4}{c|}{All Categories}                            & \multicolumn{4}{c|}{Seen Categories}                           & \multicolumn{4}{c|}{Unseen Categories}                         \\ \cline{2-13} 
\multicolumn{1}{|c|}{}                         & BLEU           & METEOR        & chrF++        & TER           & BLEU           & METEOR        & chrF++        & TER           & BLEU           & METEOR        & chrF++        & TER           \\ \hline
mBART                                          & 0.49          & 0.37          & 0.63          & 0.46          & 0.55          & 0.40          & 0.68          & 0.42          & 0.41          & 0.33          & 0.57          & 0.50          \\ \hline
mB-LAD                                         & 0.49          & \textbf{0.39} & 0.67          & 0.44          & \textbf{0.56} & \textbf{0.41} & \textbf{0.71} & \textbf{0.41} & 0.40          & 0.36          & 0.62          & 0.48          \\ \hline
mB-LAD+                                        & \textbf{0.50} & \textbf{0.39} & \textbf{0.68} & \textbf{0.43} & 0.55          & \textbf{0.41} & \textbf{0.71} & \textbf{0.41} & \textbf{0.44} & \textbf{0.38} & \textbf{0.64} & \textbf{0.46} \\ \hline
mB-LAD-SL                                      & 0.48          & \textbf{0.39} & 0.66          & 0.45          & 0.54         & \textbf{0.41} & 0.70          & \textbf{0.41} & 0.40          & 0.36          & 0.61          & 0.49          \\ \hline
\end{tabular}}
\caption{Official WebNLG20 testset results for Pretrained Multilingual models on English text.}
\label{tab:WebNLG20officialen}
\end{table*}

\subsection{Multilingual Generalisation}

Tables~\ref{tab:WebNLG20} contains full results for English and Russian on WebNLG20* respectively. We include the \textit{Word} configuration (see Section~\ref{sec:monoVmulti}), as well as \textit{Char}, \textit{BPE}, and \textit{SP}, which are variations that use characters, Byte-Pair-Encoding, and SentencePiece as subword units respectively. \textit{Copy} refers to the copy mechanism model by \citet{roberti2019copy}. 
The \textit{SP} model performs very well for seen categories, but fails to generalise on unseen data. The \textit{Copy} model performs well for unseen categories in English, but underperforms in Russian as values for it are only partially translated, i.e. some values in the MR may appear in English while others appear in Russian. This is challenging for \textit{Copy} models as the target reference does not closely match the input, but \textit{LAD} can handle it more robustly. 

Observing the output, \textit{LAD}'s main advantage is that it avoids under- and over-generating values as they are being controlled by the placeholders.\footnote{We provide output examples in the Appendix.}
\textit{SP} is often the most fluent of the models, but for longer input it tends to under-generate and miss values. The \textit{Copy} model tends to repeat values, which can be attributed to the fact that it is based on characters where long-distance dependencies are hard to maintain.
On the other hand, \textit{Copy} can potentially generate more relevant output since it can copy words from attributes as well as values. 

Overall, \textit{LAD} 
helps the multilingual model outperform all other models in both English and Russian. It is especially beneficial in generalising to unseen data, as was its main objective after all. 

\subsection{Generalising with Pretrained Models}

Here we explore the generalisation capabilities of multilingual pretrained models, by replacing the underlying NLG model with mBART \cite{liu-etal-2020-multilingual-denoising}, a multilingual denoising autoencoder pretrained on a large-scale dataset containing 25 languages (CC25). Similarly to \citet{kasner-dusek-2020-train}, we fine-tune mBART with the default EN-RO configuration for up to 10000 updates. Using mBART as the underlying model also helps facilitate a comparison against a configuration that is similar 
to many of the state of the art participants in the WebNLG 2020 Challenge, 
although some  of them used different pretrained models.

Table \ref{tab:WebNLG20_mbart} shows the performance of the fine-tuned models on the WebNLG20* dataset. The mBART-based model outperforms the non-delexicalisation \textit{SP}, and non-pretrained \textit{LAD} in English. However, \textit{LAD} still performs better in Russian. 
This makes sense as the CC25 dataset is heavily biased towards the English language and contains double the amount of tokens compared to Russian, and much more compared to other lower-resource languages.
Combining the LAD framework with mBART (\textit{mB-LAD}) resulted in a general improvement in performance, especially for lower-resource unseen data. However, as discussed in Section \ref{sec:ablation}, the VAPE component remains to some degrees susceptible to unseen contexts. To tackle this issue, we improve VAPE by pre-loading mBART and fine-tuning it for value post-editing as well (\textit{mB-LAD+}), achieving 3 and 29 points increase in BLEU score for unseen English over the vanilla \textit{mBART} and \textit{LAD} models, and 26 and 20 points for unseen Russian. 
Additionally, to take advantage of mBART's denoising ability, we extend the fine-tuned VAPE to edit the ``exact" relexicalised NLG output and 
provide a sentence-level output (\textit{mB-LAD-SPE}), i.e. edits are not exclusively focused on the values. Results show that \textit{mB-LAD-SPE} improves further \textit{mB-LAD+} on Russian in both seen and unseen.

Table \ref{tab:WebNLG20officialen} and \ref{tab:WebNLG20officialru} also shows the automatic evaluation of the fine-tuned mBART models on the official WebNLG20 Challenge testset; the official test set had no unseen subset of Russian. The results are consistent with the findings in our previous experiments, with small improvements of LAD-based mBART models over the mBART-base.

\begin{table}[!t]
\renewcommand{\arraystretch}{1.1}
\centering
\resizebox{185pt}{!}{\begin{tabular}{|l|c|c|c|c|c|c|c|c|c|c|c|c|}
\hline
\multicolumn{1}{|c|}{\multirow{2}{*}{Russian}} & \multicolumn{4}{c|}{All/Seen Categories}                       \\ \cline{2-5} 
\multicolumn{1}{|c|}{}                         & BLEU           & METEOR        & chrF++        & TER           \\ \hline
mBART                                          & 0.43          & 0.55          & 0.56          & 0.52          \\ \hline
mB-LAD                                         & 0.42          & \textbf{0.61} & \textbf{0.64} & 0.50          \\ \hline
mB-LAD+                                        & 0.38          & 0.59          & 0.61          & 0.53          \\ \hline
mB-LAD-SL                                      & \textbf{0.44} & \textbf{0.61} & 0.63          & \textbf{0.48} \\ \hline
\end{tabular}}
\caption{Official WebNLG20 testset results for Pretrained Multilingual models on Russian text.}
\label{tab:WebNLG20officialru}
\end{table}

\begin{table}[!t]
\centering
\renewcommand{\arraystretch}{1.1}
\resizebox{170pt}{!}{\begin{tabular}{|l|c|c|c|c|}
\hline
\multirow{2}{*}{Russian} & \multicolumn{4}{c|}{All Categories}                           \\ \cline{2-5} 
                         & BLEU          & METEOR        & chrF++        & TER           \\ \hline
Word                     & 0.02          & 0.16          & 0.21          & 0.95          \\ \hline
Char                     & 0.01          & 0.14          & 0.19          & 0.90          \\ \hline
BPE                      & 0.02          & 0.15          & 0.20          & 0.97          \\ \hline
SP                       & 0.02           & 0.16          & 0.20          & 0.93          \\ \hline
LAD                      & \textbf{0.04} & \textbf{0.22} & \textbf{0.26} & \textbf{0.84} \\ \hline

\end{tabular}}
\caption{WebNLG17 results for Multilingual models.}
\label{tab:WebNLG17Rus}
\end{table}

\subsection{Synthetic Data}

We use the WebNLG17 automatically translated Russian ``silver'' data, to determine how useful they are for training multilingual concept-to-text NLG. As preliminary results were not promising, we limit the scope of the experiment to only a few systems. 
Table~\ref{tab:WebNLG17Rus} gathers the results 
It is apparent that automatically translated data are insufficient; 
\textit{LAD} seems to more consistently achieve higher performance than other models, but all scores are too low to draw any sufficiently supported conclusions.

\section{Conclusion}

We proposed Language Agnostic Delexicalisation, a novel delexicalisation framework that matches and delexicalises MR values in the text independently of the language. For relexicalisation, an automatic value post editing model adapts the values to their context. 
Results 
show that multilingual models outperform monolingual models, and that 
LAD outperforms previous work in improving the performance of multilingual models, especially in low resource conditions. LAD also improves on the performance of pre-trained language models achieving state-of-the-art results. The automatic value post editing component is especially beneficial in morphologically rich languages. 



\bibliographystyle{acl_natbib}
\bibliography{anthology,acl2021}

\section{Appendices}

\appendix

\section{Configurations} 

The multilingual NLG and VAPE use a transformer as underlying architecture. We use the fairseq toolkit for our experiments \cite{ott-etal-2019-fairseq}. The models are trained with shared embeddings, 8 attention heads, 6 layers, 512 hidden size, 2048 size for the feed forward layers. We trained with 0.3 dropout, adam optimiser with a learning rate of 0.0005. The NLG are trained with early stopping and patience set to 20. Automatic value post edit models are trained with the same configuration but patience was set to 6.
For the copy mechanism-based model we use the EDA-CS 
implementation provided by \citet{roberti2019copy} with the default configuration. Due to its extremely high computational training cost, the models are trained for 15 epochs.
BPE and SentencePiece \cite{kudo-richardson-2018-sentencepiece} models are trained with a vocabulary size set to 12000 tokens.

For all models in our experiments, the input consists of a simple linearisation of the MRs. Particularly, for the delexicalisation based models, the values are extended with their respective placeholders as shown in the following example: ``ENTITY\_1 meyer werft \textit{location} ENTITY\_2 germany.

\section{Input examples} 

\begin{figure}[th]
\centering
  \includegraphics[width=\columnwidth]{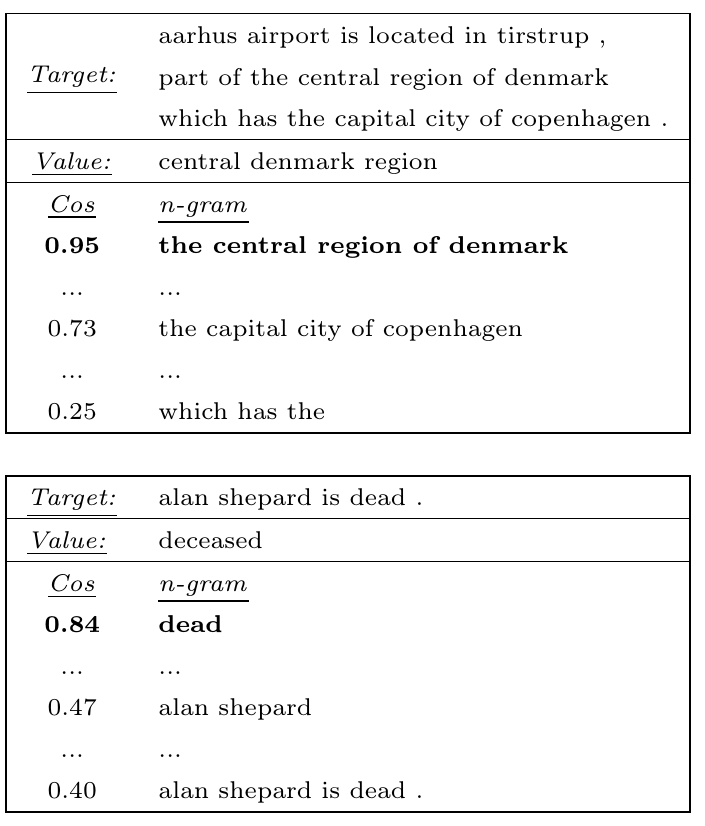}
  \caption{Examples of LAD's value mapping to target reference n-grams.}
  \label{tab:lad_delex}
\end{figure}

Figure~\ref{tab:lad_delex} shows some examples of how, during training, LAD maps MR values to n-grams of the target reference, based on the similarity of their representations. We can observe that these values could not have been matched by exact and n-gram delexicalisation as they constitute significant paraphrases of the value.

\begin{figure}[bh]
\centering
  \includegraphics[width=\columnwidth]{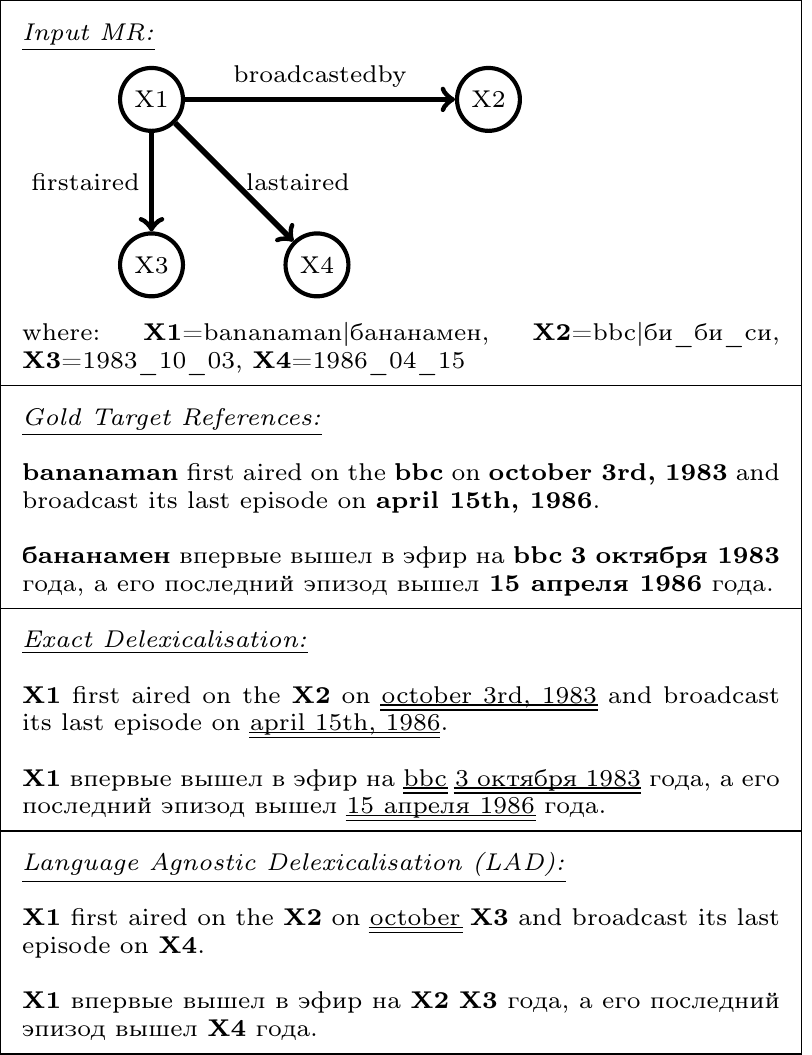}
  \caption{Delexicalisation on WebNLG Challenge 2020 with target output in English and Russian. Double underlining marks text missed by delexicalisation.}
  \label{tab:example_input}
\end{figure}

\begin{figure}[bh]
\centering
  \includegraphics[width=\columnwidth]{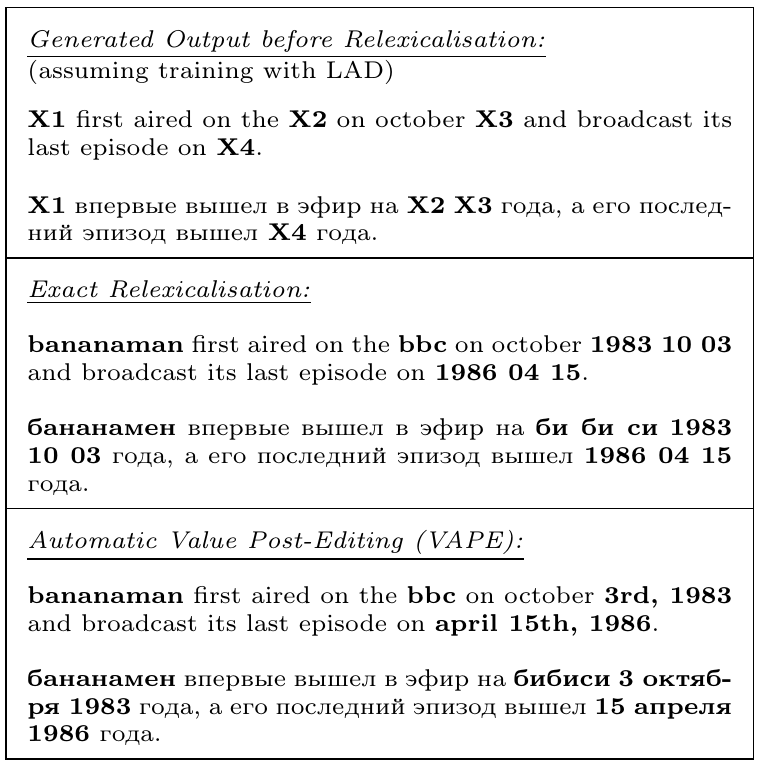}
  \caption{Relexicalisation examples; double underlining marks errors that ignore context.}
  \label{tab:example_relex}
\end{figure}

\begin{figure*}[tbh]
\centering
  \includegraphics[width=\linewidth]{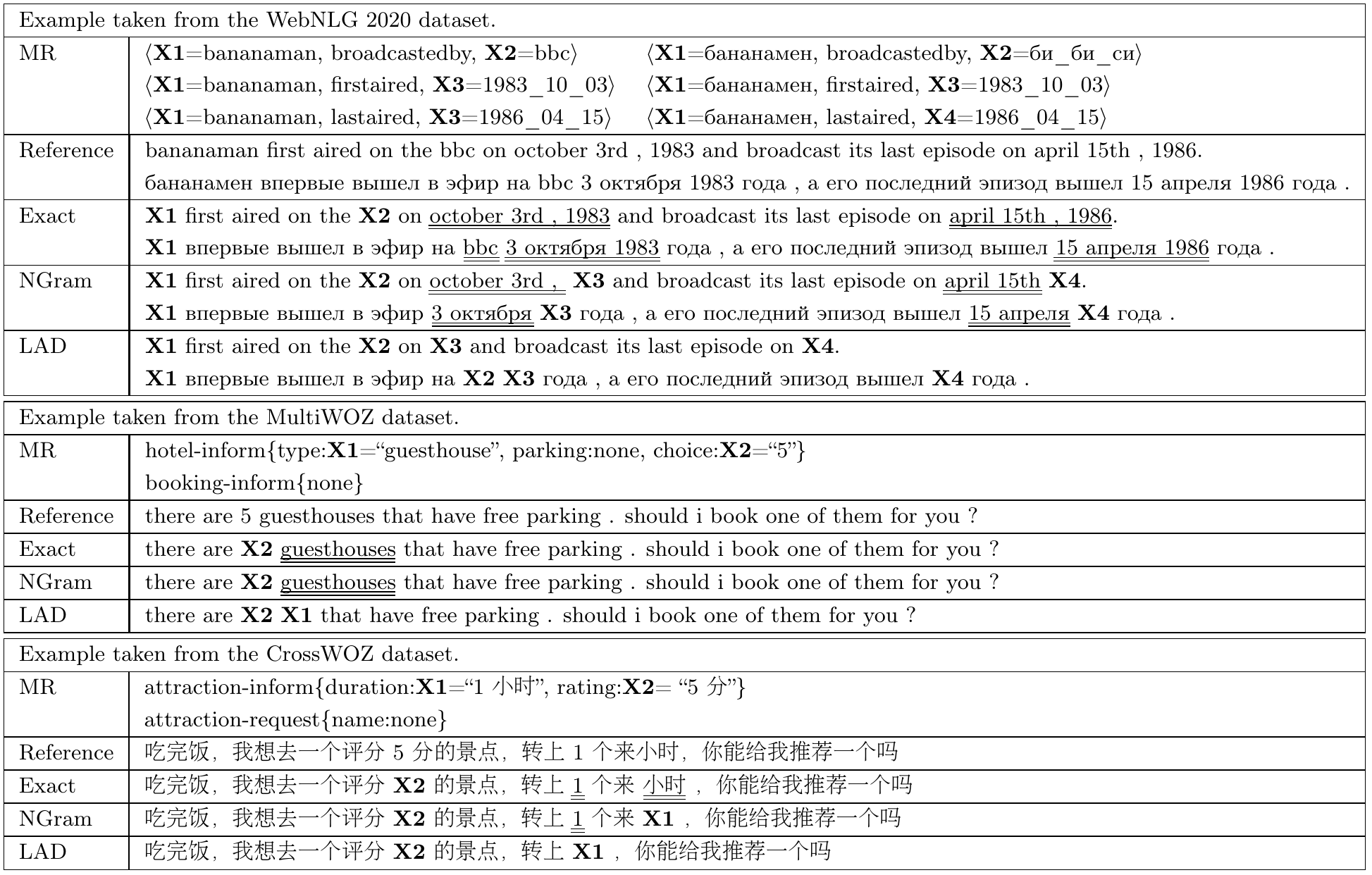}
  \captionof{table}[]{Dataset examples and delexicalisation output; double underlining marks text that was missed.}
  \label{tab:mrexamples}
\end{figure*}

Figure~\ref{tab:example_input} and \ref{tab:example_relex} show some additional examples of delexicalisation and relexialisation for the various approaches from the WebNLG Challenge 2020. 
Table~\ref{tab:mrexamples} shows more delexicalisation examples from WebNLG, MultiWOZ and CrossWOZ datasets, where we can observe the shortcomings of exact and n-gram delexicalisation.

\section{Output examples} 

\begin{figure}[th]
\centering
  \includegraphics[width=\columnwidth]{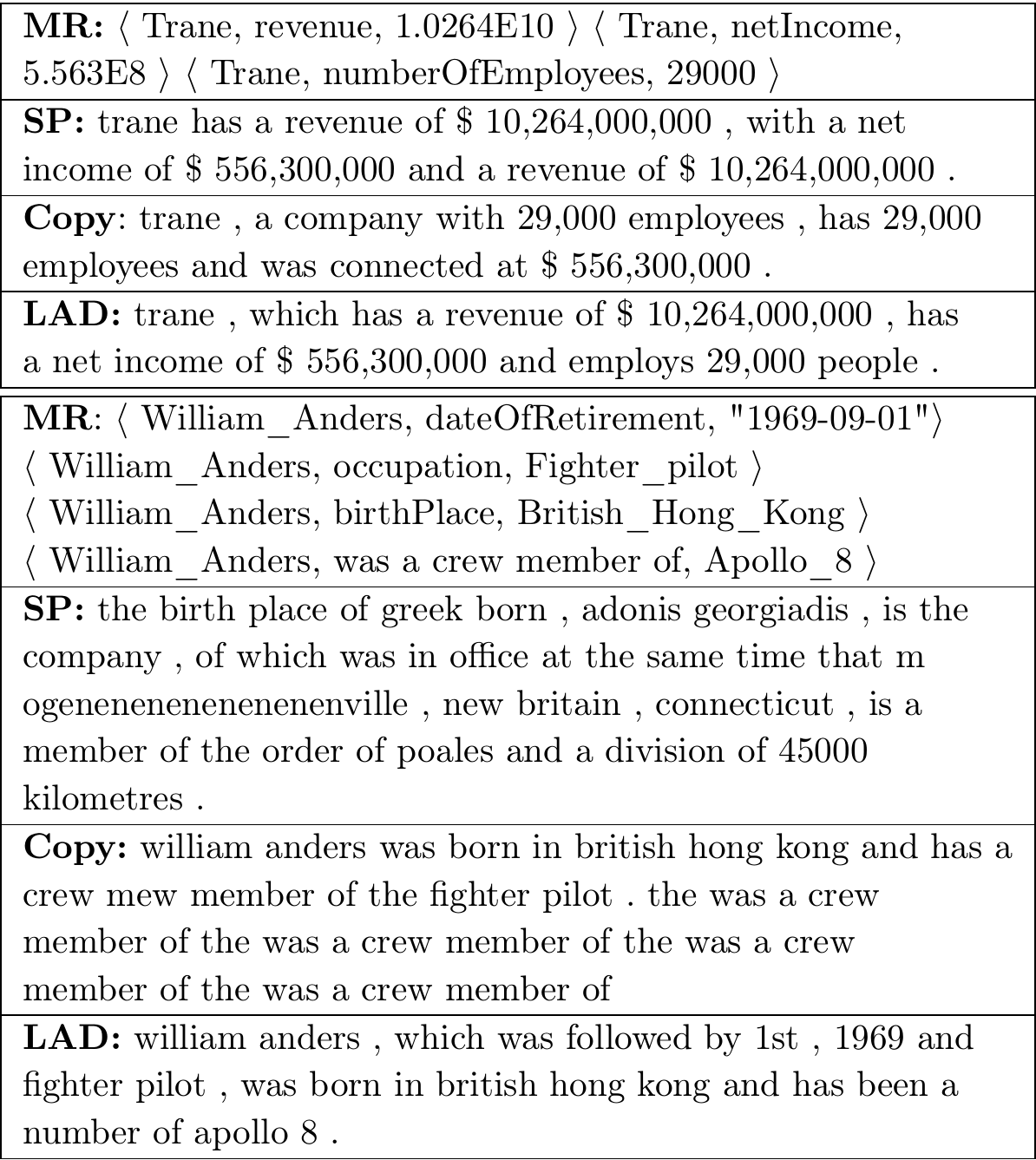}
  \captionof{table}[]{Output text from three different systems in Russian.}
  \label{tab:example_gen_rus}
\end{figure}

Table~\ref{tab:example_gen} and \ref{tab:example_gen_rus} present some examples for English and Russian output respectively. The examples include output from SentencePiece (SP), Copy, and LAD systems.

\begin{table}[bh]
\centering
  \includegraphics[width=\columnwidth]{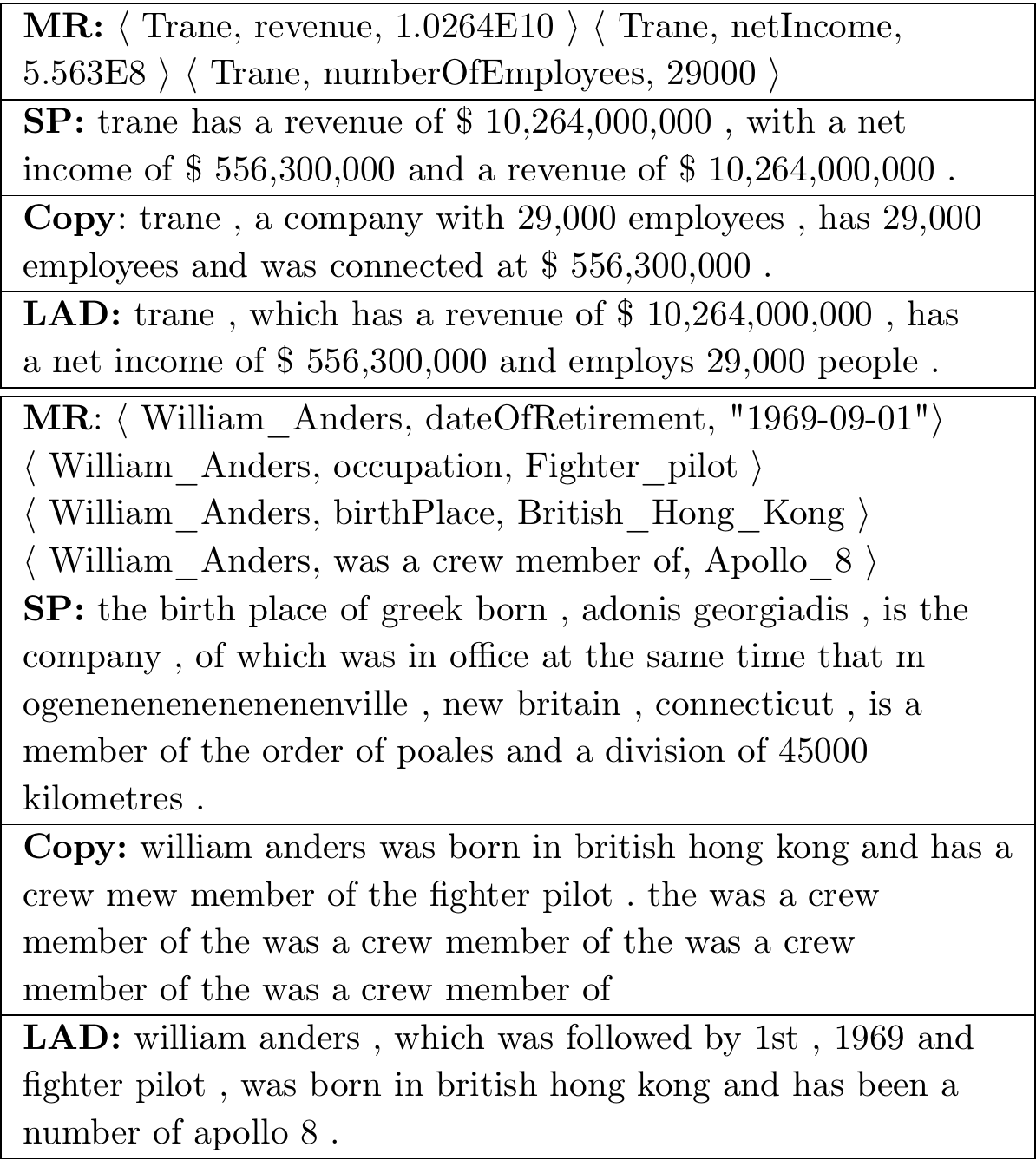}
\captionof{table}[]{Output text from three different systems in English.}\label{tab:example_gen}
\end{table}

\end{document}